\documentclass{teleai}

\usepackage{amsmath,amsfonts,bm}

\def\eqref#1{equation~\ref{#1}}
\def\1{\bm{1}}

\DeclareMathAlphabet{\mathsfit}{\encodingdefault}{\sfdefault}{m}{sl}
\SetMathAlphabet{\mathsfit}{bold}{\encodingdefault}{\sfdefault}{bx}{n}

\usepackage{amssymb}
\usepackage{booktabs,graphicx,natbib}
\usepackage{xspace}
\usepackage{wrapfig}
\usepackage{longtable}
\usepackage{multirow}

\newcommand{\method}{ReCam\xspace}
\newcommand{\correspondingmark}{\textsuperscript{\ensuremath{\dagger}}}

\hypersetup{
  pdftitle={Recast the Past: Stable Long Video Generation with Full Context Memory},
  pdfauthor={Yukuo Ma, Cong Liu, Junke Wang, Junqi Liu, Haibin Huang, Zuxuan Wu, Xuelong Li}
}

\begin{document}

\title{Recast the Past: Stable Long Video\\
Generation with Full Context Memory}

\author[1,2]{Yukuo Ma}
\author[2]{Cong Liu}
\author[1]{Junke Wang}
\author[2]{Junqi Liu}
\author[2]{Haibin Huang}
\author[1]{Zuxuan Wu\correspondingmark}
\author[2]{Xuelong Li\correspondingmark}

\affiliation[1]{Fudan University}
\affiliation[2]{Institute of Artificial Intelligence (TeleAI), China Telecom}
\contribution{\correspondingmark Corresponding authors.}
\metadata[Keywords]{Long video generation; video memory; diffusion transformers; temporal consistency}
\metadata[Github]{\url{https://github.com/scottykma/Recam}}
\date{October 16, 2026}

\abstract{
Long-horizon video generation requires retaining an ever-expanding 
history, but conditioning on its growing token sequence is costly. 
Existing methods restrict or compress historical context, potentially 
losing information needed for future generation. 
This work proposes \method, which groups local spatiotemporal blocks of 
historical latents into vectors and projects them directly into the 
high-dimensional hidden space of a Diffusion Transformer. 
The model learns to use the resulting memory tokens through native 
self-attention, reducing token count while retaining block-level 
access to the full history. To mitigate temporal drift during 
autoregressive rollout, we further introduce dual-pointer denoising with 
separate noise schedules for recent context and target frames. 
Among the evaluated methods, \method achieves the highest VBench-Long 
overall score of 80.1 and competitive static-scene revisitation 
performance on R2M-Bench. It also leads on all four consistency metrics 
for off-screen subject reappearance, including clothing consistency 
of 4.09 versus 3.03 for the best baseline.
}

\maketitle

\section{Introduction}

Advances in autoregressive video models have enabled streaming and interactive generation for applications such as 
world modeling \citep{genie3, hywp2025, dreamx2026, matrix2026, lingbotworld2026}, 
embodied intelligence \citep{dreamzero2026, cosmos-policy2026, lingbotva2026}, 
and creative content generation \citep{seedance2025, wan2025, magi2025}. 
These applications shift the focus from generating locally plausible short clips to maintaining long-term memory 
over extended temporal horizons (\textit{e.g.}, world models must maintain consistent scene states across interactions, whereas embodied agents must retain information from earlier observations to support subsequent decisions).

\begin{figure*}[ht]
\centering
\includegraphics[width=\textwidth]{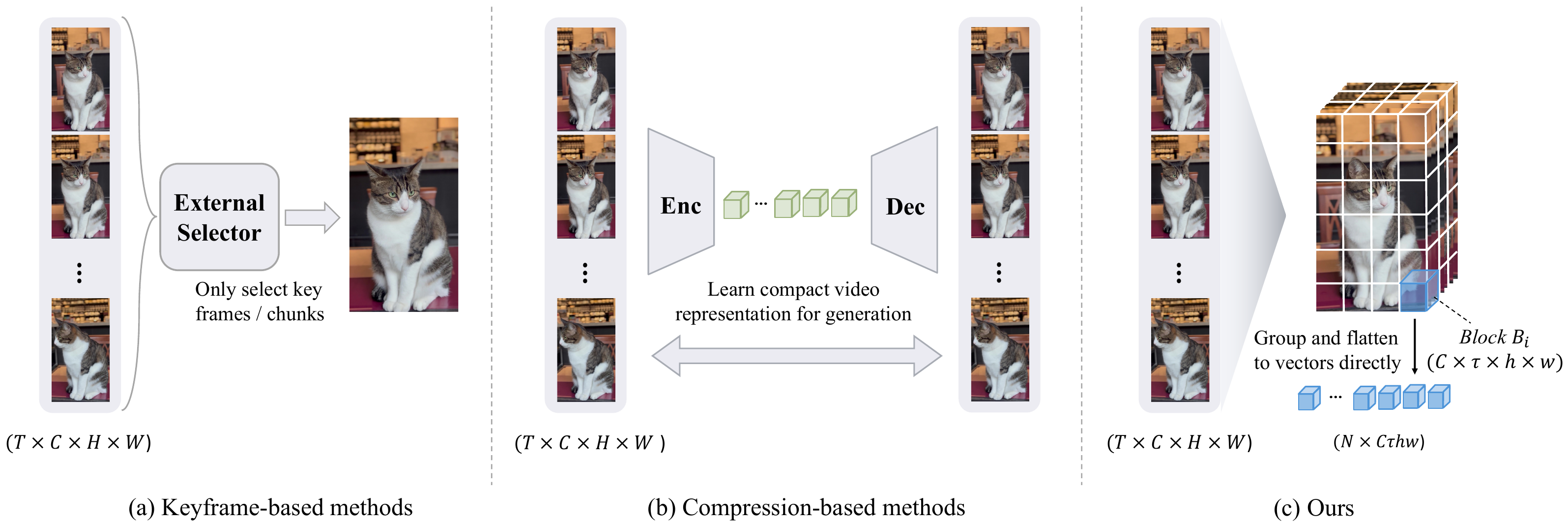}
\caption{\textbf{A coarse read on fine-grained history.}
We compare our method with other representative methods. Keyframe-based approaches, 
such as CaM \citep{cam2025}, typically employ an external rule-based selector to retrieve 
a subset of frames as memory. In contrast, compression-based methods \citep{tinyhistory2025} 
encode history into a compact representation through a separate reconstruction stage. 
Different from these approaches, \method allows the video model itself to learn how to represent 
history and how to retrieve it via its native attention mechanism. 
}
\label{fig:compare_retrieval}
\end{figure*}

Given that relevant information may vary across temporal positions and 
denoising stages \citep{memlearner2026}, effective long-term memory requires 
access to such evidence throughout the accumulated history.
However, naively retaining the complete history incurs substantial 
computational and memory costs for videos ranging from ten seconds to 
one minute, which is a practical range for these applications. 
Existing methods can be broadly categorized along two design dimensions (Fig.~\ref{fig:compare_retrieval}): how historical information
is retrieved and how it is preserved.
Selecting historical frames or sparsifying attention reduces conditioning 
cost by restricting the observations or interactions available to the 
generator \citep{cam2025,moc2026,moga2026}. 
Yet, the resulting selection may not fully capture the historical evidence 
needed, leaving relevant 
cues inaccessible even when they remain in the history.
Other methods encode history into compact embeddings or
mediate access through a limited set of query
tokens \citep{tinyhistory2025,memlearner2026}.
These designs can retain full-history coverage, yet may
discard fine-grained details needed when a scene is revisited
or a subject reappears.

These limitations motivate a method that allows video models to leverage the full history during autoregressive rollout directly. 
Our key insight is that a coarse read does not require a coarse historical representation. 
We reduce the historical token count by retrieving at a coarser granularity while retaining full-history coverage.
Each local spatiotemporal block of historical latents is flattened 
into a vector and projected into the high-dimensional hidden space of a Diffusion Transformer. The resulting memory representations are learned jointly with the video model and processed through native self-attention. This allows local detail to be encoded within each memory token, giving the model access to every historical block through a shorter token sequence.

Nevertheless, effective use of history requires more than access: errors in generated frames can propagate to subsequent predictions~\citep{exposurebias2024,framepack2025,tempomaster2026}. To mitigate this drift, we introduce dual-pointer denoising, which assigns separate noise schedules to recent context frames and current target frames. At inference, context frames start from pure noise and follow a faster denoising trajectory, providing temporal cues while reducing the influence of errors in previously generated context.

Building on these designs, we propose \method, which enables stable and consistent long-video generation. 
Our contributions are summarized as follows:
{\setlength{\emergencystretch}{2em}
\begin{itemize}
    \item \textbf{Full-context memory.} We introduce \method, a simple yet effective method that allows each token to encode a larger amount of spatiotemporal content, enabling end-to-end training for long-video generation and yielding a native long-term memory mechanism.
    \item \textbf{Dual-pointer denoising.} We assign separate denoising trajectories to recent context frames and current target frames, providing temporal cues while masking prediction errors in the context frames.
    \item \textbf{Empirical evaluation.} Extensive experiments on complementary benchmarks show that \method achieves the highest overall VBench-Long score among the evaluated methods \citep{vbench++2024}, competitive performance on static-scene revisitation \citep{r2mbench2026}, and the best results across all four metrics for off-screen subject reappearance \citep{outofsight2026,msvbench2026}. These results support the effectiveness of \method as a native long-term memory mechanism.
\end{itemize}}

\section{Method}
\label{sec:method}

\method combines a learned representation of accumulated video history with dual-pointer denoising of recent context. We first describe how historical latents are turned into memory tokens and accessed through native self-attention, then explain how recent context is incorporated during training and autoregressive rollout.

\subsection{A Coarse Read on Fine-Grained History}
\label{sec:history_recasting}

We reduce the number of historical tokens by grouping local spatiotemporal content into larger read units. Each unit is represented in the high-dimensional hidden space of the DiT, allowing the model to learn how to encode and use local detail while retaining access to the full history.

\paragraph{Recasting the history.}
Let $F_H \in \mathbb{R}^{C \times T \times H \times W}$ denote the $T$ accumulated historical latent frames. We partition $F_H$ into local spatiotemporal blocks $\{B_i\}_{i=1}^{N_b}$, where $B_i \in \mathbb{R}^{C \times \tau \times h \times w}$. Each block is flattened into a vector $\operatorname{vec}(B_i) \in \mathbb{R}^{C\tau h w}$ and projected by a lightweight MLP $f_{\phi}$:
\begin{equation}
    m_i = f_{\phi}\!\left(\operatorname{vec}(B_i)\right),
    \qquad m_i \in \mathbb{R}^{d}.
\end{equation}
Here, $d$ is the hidden dimension of the DiT. The default block shape is $(C,\tau,h,w)=(16,2,8,8)$, which packs $128$ spatiotemporal positions into a $2048$-dimensional vector. This is smaller than the hidden dimension of a large DiT, such as $5120$ in Wan-14B. 
The projection therefore avoids imposing an additional low-dimensional memory bottleneck.

\begin{figure}[!t]
    \centering
    \includegraphics[width=0.9\linewidth]{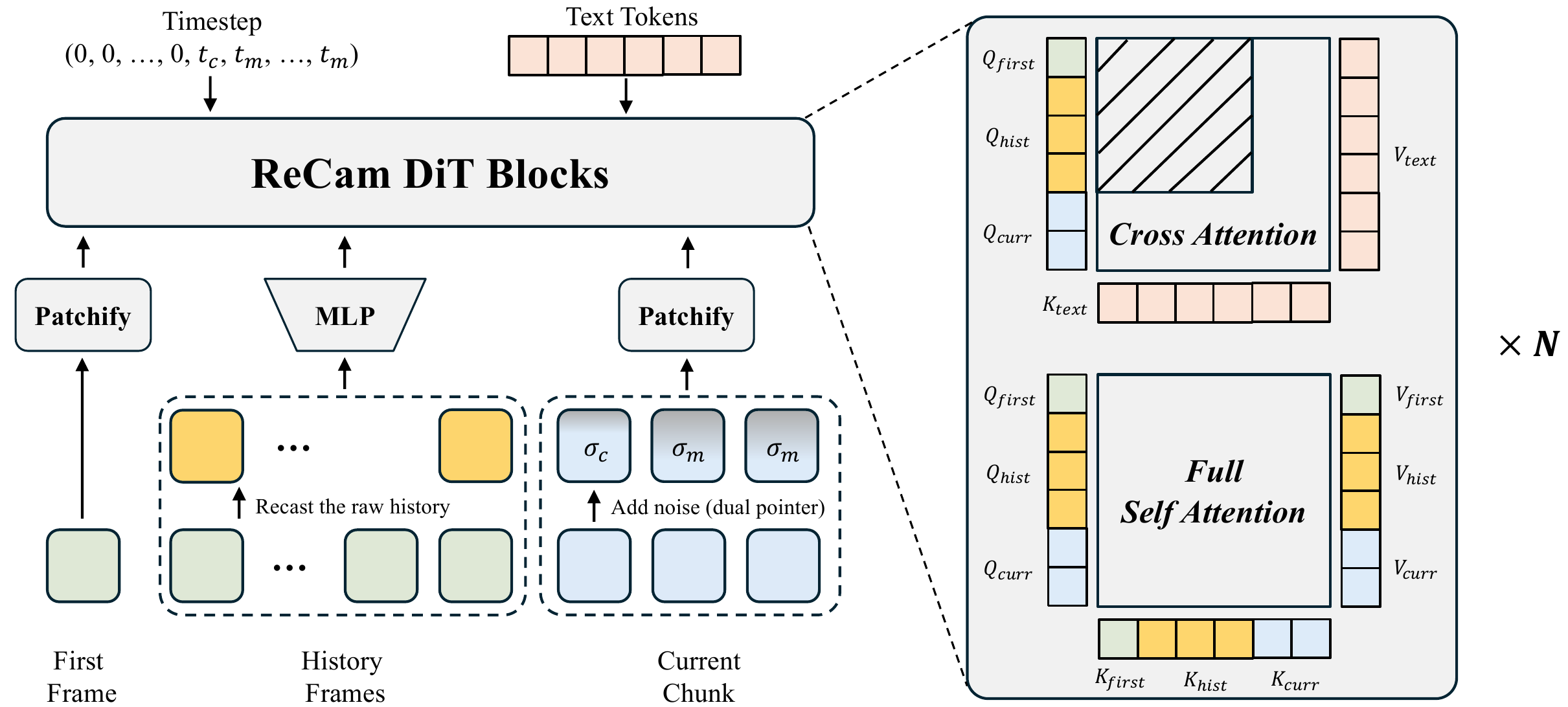}
    \caption{\textbf{Overview of \method.}
Historical latents are grouped into local spatiotemporal blocks and projected into memory tokens by a lightweight MLP. The first frame remains a clean visual anchor. Native self-attention connects the historical memory and current chunk, while text cross-attention is applied only to the current chunk.}
    \label{fig:architecture_overview}
\end{figure}

\paragraph{Attention with memory tokens.}
As shown in Fig.~\ref{fig:architecture_overview}, memory tokens are concatenated in temporal order before the current-chunk tokens and processed through native DiT self-attention. We assign 3D RoPE coordinates to preserve the spatiotemporal ordering of the packed history. Memory tokens use negative temporal coordinates; native DiT tokens, including the noisy context, retain their original positional encodings with temporal coordinates starting from zero.

Coordinates are defined on the native DiT patch grid, whose patch size is $(1,2,2)$ in latent space. For a block indexed by $(n,i,j)$ along the temporal, height, and width dimensions, we use
\begin{equation}
    \pi_{n,i,j}
    =
    \left(-T+n\tau,\; i\frac{h}{2},\; j\frac{w}{2}\right),
\end{equation}
where the spatial coordinates match the first native patch in the block. 
With the historical memory already determined, text cross-attention is applied only to the current-chunk tokens.

\subsection{Dual-Pointer Denoising for Stable Rollout}
\label{sec:dp_denoising}

Alongside long-term memory, we retain recent frames as short-term context to provide appearance and motion cues. However, reusing generated context can propagate prediction errors across chunks. Dual-pointer denoising mitigates this drift while retaining useful temporal guidance.

\paragraph{Separate noise schedules.}
Reusing fully clean generated context can amplify prediction errors during rollout~\citep{hiar2026}. We keep the original first frame as a clean visual anchor and instead provide a noisy view of the recent context at each denoising step:
\begin{equation}
    \tilde O^{(k)}
    =
    \left(1-\sigma_{\mathrm c}^{(k)}\right)O
    +\sigma_{\mathrm c}^{(k)}\epsilon,
    \qquad
    \epsilon \sim \mathcal{N}(\mathbf{0},\mathbf{I}),
\end{equation}
where $O$ denotes the recent context. Context and current-chunk noise levels follow the same shifted-linear schedule family~\citep{sd2024}, but use different shifts. Given the base schedule $\bar{\sigma}^{(k)}=1-k/K$ over $K$ denoising steps, we set
\begin{equation}
    \sigma_{j}^{(k)}
    =
    \frac{s_{j}\bar{\sigma}^{(k)}}
    {1+\left(s_{j}-1\right)\bar{\sigma}^{(k)}},
    \qquad j\in\{\mathrm{c},\mathrm{m}\},
\end{equation}
where $s_{\mathrm c}$ and $s_{\mathrm m}$ control the context and current chunk, respectively. Setting $s_{\mathrm c}\leq s_{\mathrm m}$ ensures $\sigma_{\mathrm c}^{(k)}\leq\sigma_{\mathrm m}^{(k)}$, so the context is no noisier than the current chunk and can guide its generation. With the default $s_{\mathrm c}=2$, the context starts from pure noise and remains noisy through much of the schedule, limiting the influence of errors in the reused frames. The schedule restores the clean context as denoising ends.

\paragraph{An input-level shortcut for noisy context.}
Noisy context introduces additional denoising states and can make optimization harder~\citep{helios2026}. We expose the context denoising direction through paired noisy views, allowing the model to focus its prediction objective on the current chunk. For clean context $O$, the flow-matching target is $v_{\mathrm{ctx}}^{*}=\epsilon-O$, where $\epsilon\sim\mathcal{N}(\mathbf{0},\mathbf{I})$. During training, we construct two views with shared noise at levels $\sigma_{\mathrm c}\leq\sigma_{\mathrm m}$:
\begin{equation}
    \tilde O_{\mathrm c}
    =
    \left(1-\sigma_{\mathrm c}\right)O
    +\sigma_{\mathrm c}\epsilon,
    \qquad
    \tilde O_{\mathrm m}
    =
    \left(1-\sigma_{\mathrm m}\right)O
    +\sigma_{\mathrm m}\epsilon.
\end{equation}
Under the linear flow path, these two pointers satisfy
\begin{equation}
    \tilde O_{\mathrm m}-\tilde O_{\mathrm c}
    =
    \left(\sigma_{\mathrm m}-\sigma_{\mathrm c}\right)
    \left(\epsilon-O\right)
    =
    \left(\sigma_{\mathrm m}-\sigma_{\mathrm c}\right)
    v_{\mathrm{ctx}}^{*}.
\end{equation}
When the noise levels differ, this relation allows the context denoising direction to be inferred from the 
paired views and their noise levels.
It provides an input-level shortcut aligned with the $v$-prediction parameterization in flow matching: 
the model can directly access the context velocity and focus on denoising the 
current chunk, without having to learn to reconstruct $O$. 
At inference, the model also does not try to reconstruct a clean $O$; 
instead, it directly consumes the noisy context, 
thereby circumventing the accumulated generation errors that a clean 
$O$ would contain.
We concatenate both views and their noise levels along the channel 
dimension and feed them into the DiT. The $\sigma_{\mathrm m}$ channel 
is optional because the conditioning timestep $t_{\mathrm m}$ already 
encodes its value. 
No additional prediction loss is imposed on the context $\tilde O_{\mathrm c}$.

\subsection{Training}

We train the model to denoise the current chunk conditioned on the 
preceding video history. Let $z_0$ denote the current chunk and 
$\mathcal{H}$ the conditioning inputs from Secs.~\ref{sec:history_recasting} and~\ref{sec:dp_denoising}. 
We construct $z_{\sigma}=(1-\sigma)z_0+\sigma\epsilon$, 
with $\epsilon\sim\mathcal{N}(\mathbf{0},\mathbf{I})$, and optimize the 
standard flow-matching objective:
\begin{equation}
    \mathcal{L}_{\mathrm{FM}}
    =
    \mathbb{E}
    \left[
    \left\|
    v_{\theta}(z_{\sigma},\sigma\mid\mathcal{H})
    -
    (\epsilon-z_0)
    \right\|_2^2
    \right].
\end{equation}
The loss is computed only over the current chunk. Historical tokens 
and auxiliary context channels serve as conditioning inputs.
For each sample, we randomly select a generation window and a 
history length. The immediately preceding segment is encoded into 
latent space as historical context. We progressively increase the 
maximum history length from 10 to 30 seconds. The first frame remains a 
clean anchor, while the remaining historical latents are converted 
into memory tokens and fed directly into the DiT, bypassing native 
patchification.
During training, the first latent frame of the current chunk is 
reserved as short-term context for dual-pointer denoising. At inference, 
the last latent frame of the previous chunk serves this role, 
creating a one-latent-frame overlap between consecutive chunks.

\paragraph{Optional camera control.}
For camera-controlled tasks such as scene revisitation, we follow 
\citet{mosaicmem2026} and inject camera trajectories through 
fused-PRoPE. We apply PRoPE~\citep{prope2026} after the original 
DiT RoPE and before self-attention, without adding learnable 
parameters. During training and inference, only the trajectory of 
the current generation window is provided, expressed as relative 
poses with respect to its first frame. Historical consistency must 
therefore be learned without an explicit geometry-based memory.
Please see Appendix \ref{app:camera-control} for more details.
\section{Experiments}
\label{sec:experiments}

\subsection{Experimental Setup}

We initialize \method from the high-noise expert of Wan2.2-I2V-A14B and generate 480p videos at 16 fps, adding 40 pixel frames per autoregressive step. Training uses a global batch size of 64 and three stages of 10,000 steps, with maximum history lengths of 10, 20, and 30 seconds. We use AdamW with learning rates of $10^{-5}$, $10^{-6}$, and $10^{-6}$ across the three stages. Default inference uses 40 Euler steps and classifier-free guidance.

\paragraph{Training data.}
We train on an internal corpus of 2 million general-purpose videos lasting 5--60 seconds, mostly sourced from the web and covering diverse real-world scenarios. No manually constructed memory labels, such as revisit segments identified by FOV overlap, are used. For static-scene revisitation only, we additionally train on SpatialVID-HQ~\citep{spatialvid2026} to acquire camera control. Appendix~\ref{app:training-data} provides further details.

\paragraph{Evaluation settings.}
We evaluate overall long-video quality and two forms of historical recall. The protocols and metric definitions are detailed in Appendix~\ref{app:evaluation-details}.

(1) \emph{Long-horizon generation.} The official VBench-Long I2V suite~\citep{vbench++2024} evaluates 30-second videos at 480p across six dimensions of consistency, motion, and visual quality.

(2) \emph{Static-scene revisitation.} We use 100 reference scenes from R2M-Bench~\citep{r2mbench2026} to test whether previously observed content is recovered when the camera returns. MemoryGain (MG) and the Normalized Memory Ratio (NMR) calibrate revisit consistency against controls from the same rollout across appearance fidelity, scene identity, and object identity.

(3) \emph{Off-screen subject reappearance.} Following the MSVBench protocol~\citep{msvbench2026}, we evaluate 64 histories covering 40 subjects and 10 backgrounds. Each history ends after the subject leaves the frame, and generation must recover that subject upon reappearance. In half of the histories, the subject first appears after the initial frame; in every history, it is absent for the final five seconds. We measure face, character, background, and clothes consistency against the reference history.

\begin{table*}[t]
    \centering
    \caption{
    \textbf{VBench-Long evaluation results.} We report the VBench-Long metrics on videos generated at 480p, 30 seconds. Higher values are better for all dimensions.
    }
    \label{tab:vbench-long}
    \smallskip
    \scriptsize
    \setlength{\tabcolsep}{2.5pt}
    \renewcommand{\arraystretch}{1.05}
    \resizebox{\textwidth}{!}{%
    \begin{tabular}{@{}l r *{7}{r}@{}}
        \toprule
        \textbf{Method}
        & \textbf{\#Params}
        & \shortstack{\textbf{Total} \\ \textbf{Score}}
        & \shortstack{\textbf{Subject} \\ \textbf{Consistency}}
        & \shortstack{\textbf{Background} \\ \textbf{Consistency}}
        & \shortstack{\textbf{Motion} \\ \textbf{Smoothness}}
        & \shortstack{\textbf{Dynamic} \\ \textbf{Degree}}
        & \shortstack{\textbf{Imaging} \\ \textbf{Quality}}
        & \shortstack{\textbf{Aesthetic} \\ \textbf{Quality}} \\

        \midrule
        MAGI-1~\citep{magi2025} & 24B & 78.5 & 98.2 & 97.2 & 99.4 & 21.4 & 66.4 & 55.9 \\
        FramePack~\citep{framepack2025} & 13B & 79.5 & 98.7 & 99.2 & 99.5 & 16.8 & 70.9 & 61.3 \\
        SkyReels-V2~\citep{skyreelsv2_2025} & 14B & 79.2 & 96.0 & 96.0 & 99.0 & 53.2 & 64.8 & 56.2 \\
        MMPL~\citep{mmpl2025} & 14B & 78.8 & 96.2 & 95.3 & 98.8 & 49.2 & 66.4 & 55.8 \\
        LongLive 2.0~\citep{longlive2026} & 5B & 76.5 & 92.3 & 92.5 & 99.1 & 25.1 & 70.0 & 59.8 \\
        Helios~\citep{helios2026} & 14B & 78.6 & 97.6 & 97.3 & 99.1 & 28.6 & 66.4 & 59.3 \\
        \midrule
        \method (Ours) & 14B & 80.1 & 98.1 & 97.2 & 98.5 & 36.4 & 70.0 & 62.5 \\
        \bottomrule
    \end{tabular}%
    }
\end{table*}

\begin{table*}[t]
    \centering
    \caption{\textbf{Static-scene revisitation on R2M-Bench.}
    MemoryGain (MG, top) and Normalized Memory Ratio (NMR, bottom);
    higher is better for both.}
    \label{tab:static-scene-r2m}
    \smallskip
    \scriptsize
    \setlength{\tabcolsep}{2.5pt}
    \renewcommand{\arraystretch}{1.05}
    \resizebox{\textwidth}{!}{%
    \begin{tabular}{@{}l r *{8}{r}@{}}
        \toprule
        & & \multicolumn{3}{c}{\textbf{Appearance Fidelity}}
        & \multicolumn{3}{c}{\textbf{Scene Identity}}
        & \multicolumn{2}{c}{\textbf{Object Identity}} \\
        \cmidrule(lr){3-5} \cmidrule(lr){6-8} \cmidrule(lr){9-10}
        \textbf{Method} & \textbf{\#Params}
        & \textbf{PSNR} & \textbf{SSIM} & \textbf{LPIPS}
        & \textbf{DINO-sim} & \textbf{BoQ-sim} & \textbf{MVPR-sim}
        & \textbf{App. Persist} & \textbf{Sem. Persist} \\
        \midrule
        HY-WorldPlay \citep{hywp2025} & 8B
        & 2.626 & 0.060 & 0.158 & 0.287 & 0.340 & 0.157
        & 0.096 & 0.026 \\
        Sana-WM \citep{sanawm2026} & 2.6B
        & 0.534 & 0.014 & 0.022 & 0.039 & 0.019 & 0.012
        & 0.004 & 0.002 \\
        LingBot-World-Fast \citep{lingbotworld2026} & 18B
        & 1.021 & 0.028 & 0.057 & 0.194 & 0.207 & 0.162
        & 0.069 & 0.019 \\
        LingBot-World-2-Fast \citep{lingbotworldv2-2026} & 18B
        & 1.349 & 0.037 & 0.095 & 0.216 & 0.232 & 0.117
        & 0.064 & 0.014 \\
        \midrule
        \method (Ours) & 14B
        & 1.691 & 0.045 & 0.104 & 0.286 & 0.288 & 0.212 & 0.087 & 0.026 \\
        \bottomrule
    \end{tabular}%
    }

    \par\smallskip
    \resizebox{\textwidth}{!}{%
    \begin{tabular}{@{}l r *{8}{r}@{}}
        \toprule
        & & \multicolumn{3}{c}{\textbf{Appearance Fidelity}}
        & \multicolumn{3}{c}{\textbf{Scene Identity}}
        & \multicolumn{2}{c}{\textbf{Object Identity}} \\
        \cmidrule(lr){3-5} \cmidrule(lr){6-8} \cmidrule(lr){9-10}
        \textbf{Method} & \textbf{\#Params}
        & \textbf{PSNR} & \textbf{SSIM} & \textbf{LPIPS}
        & \textbf{DINO-sim} & \textbf{BoQ-sim} & \textbf{MVPR-sim}
        & \textbf{App. Persist} & \textbf{Sem. Persist} \\
        \midrule
        HY-WorldPlay \citep{hywp2025} & 8B
        & 0.618 & 0.452 & 0.576 & 0.896 & 0.800 & 0.877
        & 0.711 & 0.897 \\
        Sana-WM \citep{sanawm2026} & 2.6B
        & 0.151 & 0.116 & 0.096 & 0.120 & 0.040 & 0.041
        & 0.025 & 0.034 \\
        LingBot-World-Fast \citep{lingbotworld2026} & 18B
        & 0.225 & 0.202 & 0.230 & 0.451 & 0.388 & 0.522
        & 0.414 & 0.563 \\
        LingBot-World-2-Fast \citep{lingbotworldv2-2026} & 18B
        & 0.380 & 0.322 & 0.404 & 0.588 & 0.505 & 0.559
        & 0.427 & 0.475 \\
        \midrule
        \method (Ours) & 14B
        & 0.457 & 0.493 & 0.444 & 0.653 & 0.518 & 0.666 & 0.468 & 0.634 \\
        \bottomrule
    \end{tabular}%
    }
\end{table*}

% Values transcribed from the supplied screenshots; no pooled means recomputed.
\begin{table}[t]
    \centering
    \caption{\textbf{Off-screen subject reappearance.} For each consistency
    metric, we report the mean metric over all cases (All), the cases where the
    subject is visible in the first frame (Present), and the cases where it
    is absent from the first frame (Absent). Bold indicates the best result
    in each column.}
    \label{tab:offscreen-subject}
    \smallskip
    \scriptsize
    \setlength{\tabcolsep}{2.5pt}
    \renewcommand{\arraystretch}{1.05}
    \resizebox{\linewidth}{!}{%
    \begin{tabular}{@{}l *{12}{r}@{}}
        \toprule
        \textbf{Method}
        & \multicolumn{3}{c}{\textbf{Face} $\downarrow$}
        & \multicolumn{3}{c}{\textbf{Character} $\downarrow$}
        & \multicolumn{3}{c}{\textbf{Background} $\downarrow$}
        & \multicolumn{3}{c}{\textbf{Clothes} $\uparrow$} \\
        \cmidrule(lr){2-4}\cmidrule(lr){5-7}\cmidrule(lr){8-10}\cmidrule(l){11-13}
        & \textbf{All} & \textbf{Present} & \textbf{Absent}
        & \textbf{All} & \textbf{Present} & \textbf{Absent}
        & \textbf{All} & \textbf{Present} & \textbf{Absent}
        & \textbf{All} & \textbf{Present} & \textbf{Absent} \\
        \midrule
        FramePack-F1~\citep{framepack2025}
        & 0.578 & \textbf{0.400} & 0.745 & 12.54 & \textbf{9.82} & 14.70 & 0.043 & 0.028 & 0.054 & 2.44 & 3.95 & 1.25 \\
        Helios~\citep{helios2026}
        & 0.748 & 0.657 & 0.771 & 13.91 & 12.64 & 14.52 & 0.063 & 0.105 & 0.043 & 1.82 & 2.69 & 1.41 \\
        LongLive 2.0~\citep{longlive2026}
        & 0.626 & 0.553 & 0.710 & 13.18 & 13.03 & 13.34 & 0.239 & 0.254 & 0.219 & 2.25 & 2.73 & 1.73 \\
        Sana-WM Streaming~\citep{sanawm2026}
        & 0.642 & 0.624 & 0.682 & 11.67 & 10.86 & 12.92 & 0.073 & 0.069 & 0.078 & 2.39 & 2.84 & 1.73 \\
        LingBot-World-Fast~\citep{lingbotworld2026}
        & 0.591 & 0.578 & \textbf{0.606} & 10.84 & 10.83 & 10.86 & 0.079 & 0.071 & 0.085 & 3.03 & 3.03 & 3.04 \\
        LingBot-World-2-Fast~\citep{lingbotworldv2-2026}
        & 0.654 & 0.601 & 0.703 & 12.07 & 11.30 & 12.89 & 0.114 & 0.075 & 0.130 & 1.58 & 1.69 & 1.47 \\
        \midrule
        \method (Ours)
        & \textbf{0.561} & 0.499 & 0.644 & \textbf{9.90} & 9.90 & \textbf{9.91} & \textbf{0.028} & \textbf{0.018} & \textbf{0.042} & \textbf{4.09} & \textbf{4.50} & \textbf{3.57} \\
        \bottomrule
    \end{tabular}%
    }
\end{table}

\subsection{Comparison with Existing Methods}

\paragraph{Long-horizon generation.}
All methods generate 30-second videos at 480p under the VBench-Long I2V protocol, except Helios, which uses its training resolution of 368p because performance degrades at other resolutions. In Tab.~\ref{tab:vbench-long}, \method achieves the highest overall score of 80.1 among the evaluated methods, supporting its effectiveness across visual quality and long-term consistency.

\begin{figure}[!htbp]
    \centering
    \includegraphics[width=0.98\linewidth]{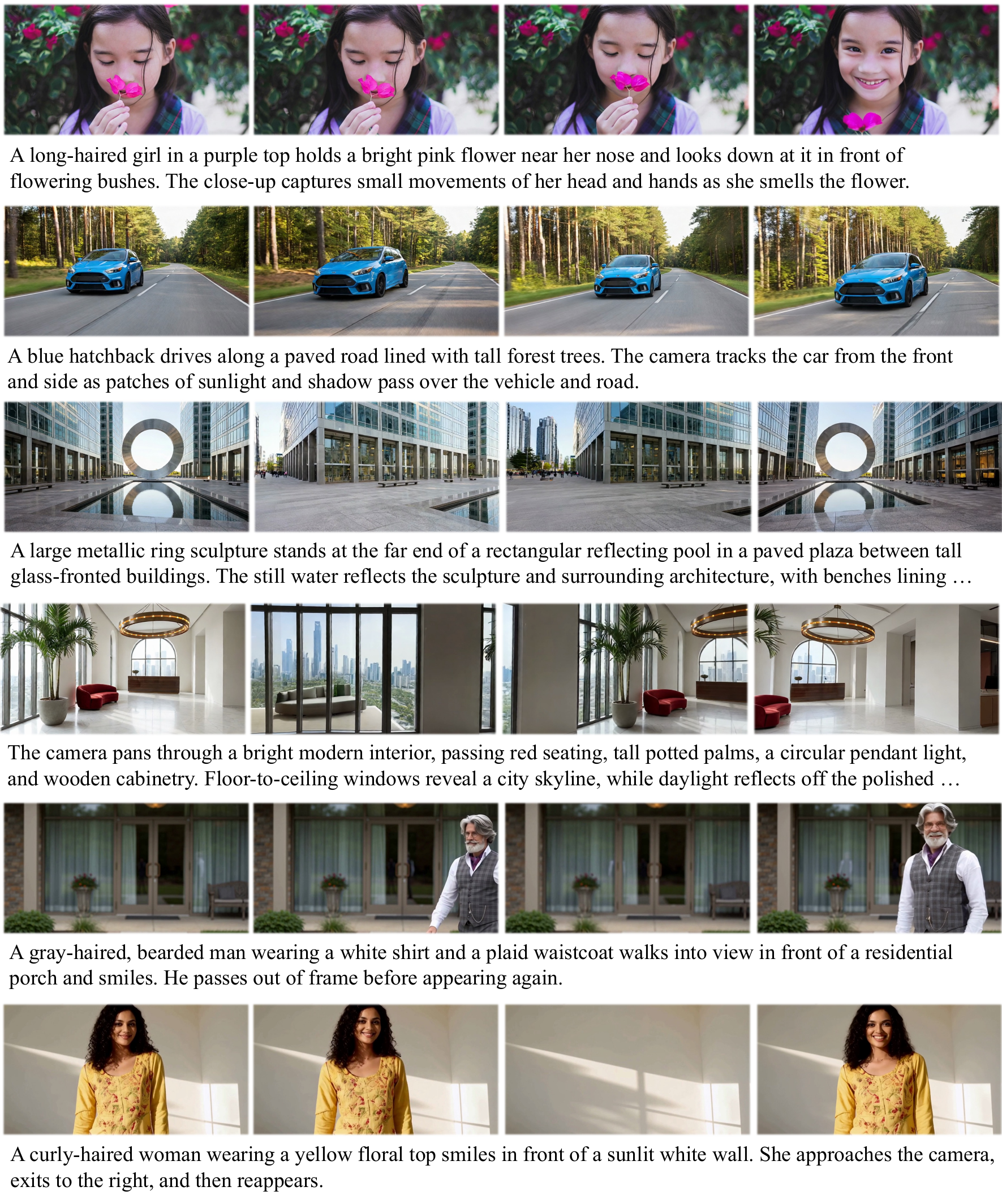}
    \caption{\textbf{Qualitative results.} From top to bottom: long-horizon generation, static-scene revisitation, and off-screen subject reappearance.}
    \label{fig:qualitative_long_video}
\end{figure}

\paragraph{Static-scene revisitation.}
Table~\ref{tab:static-scene-r2m} compares \method with raw historical KV-cache models (LingBot-World-Fast and LingBot-World-2-Fast~\citep{lingbotworld2026,lingbotworldv2-2026}), FOV-based retrieval (HY-WorldPlay~\citep{hywp2025}), and recurrent state-space memory (Sana-WM~\citep{sanawm2026}) under the same R2M-Bench protocol. Without an explicit geometry-based memory or manually constructed revisit supervision, \method remains competitive: it achieves the highest MG on MVPR-sim (0.212), ties for the highest MG on Sem.Persist (0.026), and obtains the highest NMR on SSIM (0.493).

The remaining gap is most visible against HY-WorldPlay on NMR. We observe minor viewpoint and position shifts, motion instability, and occasional degradation on second or third revisits. Limited camera-control and revisit data may contribute to these failures. The results show that learned historical memory transfers to scene revisitation, while leaving room for improvement in camera accuracy and repeated-return consistency.

\paragraph{Off-screen subject reappearance.}
We compare \method with six baselines using identical histories and reappearance targets (Tab.~\ref{tab:offscreen-subject}). FramePack supports only 19 context latents, corresponding to less than five seconds, so we use its official FramePack-F1 continuation strategy, which retains the first frame and a sliding window of recent frames.

\method leads on all four metrics over the full evaluation set, including clothes 
consistency of 4.09 versus 3.03 for the best baseline. The split results reveal 
where historical access matters most. When the subject is absent from the first 
frame, \method leads on character, background, and clothes consistency, while 
LingBot-World-Fast leads on face consistency. FramePack-F1 has a slight advantage on 
face and character consistency when the subject is initially visible, but its face 
distance increases from 0.400 to 0.745 when the subject is initially absent, 
exceeding the 0.68 same-identity threshold for the VGG-face cosine distance we use.
In the initially absent cases, identity evidence lies outside both the first frame and the recent context. Full-history coverage keeps these intermediate observations available at generation time. The results therefore support retaining access to temporally distant subject information instead of relying only on fixed anchor frames or recent observations.

% Keep the ablation tables, plots, and qualitative comparison in one float.
\begin{table}[!t]
    \centering
    \caption{\textbf{Ablation studies.} (a) History block shape and MLP
    bottleneck dimension. (b) Context design: VBench changes between the
    first and last two-second windows ($\Delta=\mathrm{last}-\mathrm{first}$);
    negative values indicate degradation. Bold marks the best result per column.}
    \label{tab:ablation-summary}
    \begingroup
    \fontsize{8}{9}\selectfont
    \setlength{\tabcolsep}{1.5pt}
    \renewcommand{\arraystretch}{1.05}
    \smallskip
    \begin{minipage}[t]{0.49\linewidth}
        \centering
        \textbf{(a) History granularity}\label{tab:history-granularity}\par\smallskip
            \begin{tabular}{@{}lrrrr@{}}
        \toprule
        \shortstack[l]{\textbf{Shape}\\$(C,\tau,h,w)$}
        & \shortstack{\textbf{MLP}\\\textbf{dim.}}
        & \textbf{VBench} $\uparrow$ & \textbf{Face} $\downarrow$ & \textbf{Character} $\downarrow$
        \\
        \midrule
        $(16,2,4,4)$ & 512
        & 79.36 & \textbf{0.577} & 10.165 \\
        $(16,2,8,8)$ & 2048
        & 80.01 & 0.695 & \textbf{9.918} \\
        $(16,4,8,8)$ & 4096
        & \textbf{80.60} & 0.662 & 10.352 \\
        $(16,16,4,4)$ & 4096
        & 80.50 & 0.650 & 10.238 \\
        \midrule
        $(16,2,8,8)$ & 128
        & 78.92 & 0.633 & 10.928 \\
        \bottomrule
    \end{tabular}

    \end{minipage}\hfill
    \begin{minipage}[t]{0.49\linewidth}
        \centering
        \textbf{(b) Context design and temporal drift}\label{tab:context-drift}\par\smallskip
            \begin{tabular}{@{}lrrrr@{}}
        \toprule
        \textbf{Context}
        & \shortstack{\textbf{Subject}\\$\Delta\uparrow$}
        & \shortstack{\textbf{Background}\\$\Delta\uparrow$}
        & \shortstack{\textbf{Imaging}\\$\Delta\uparrow$}
        & \shortstack{\textbf{Aesthetic}\\$\Delta\uparrow$} \\
        \midrule
        Clean
        & $+$0.0010 & $+$0.0009 & $-$0.0100 & $-$0.0367 \\
        Single
        & $-$0.0062 & $-$0.0027 & $-$0.0093 & $-$0.0422 \\
        Dual
        & $+$\textbf{0.0015} & $+$\textbf{0.0011}
        & $-$\textbf{0.0044} & $-$\textbf{0.0315} \\
        \bottomrule
    \end{tabular}

    \end{minipage}
    \endgroup
    \medskip
    \begin{minipage}[b]{0.635\linewidth}
        \centering
        \includegraphics[width=\linewidth]{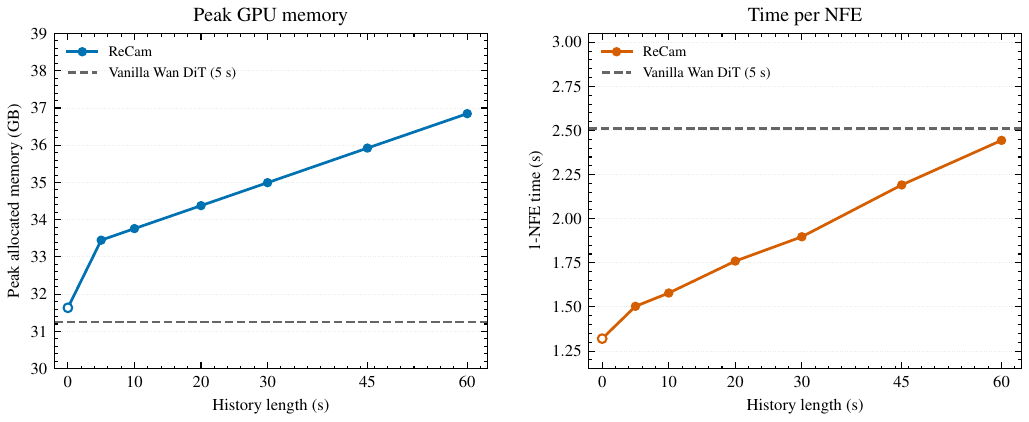}\par
        {\small\textbf{(a) History-length scaling}}
    \end{minipage}\hfill
    \begin{minipage}[b]{0.355\linewidth}
        \centering
        \includegraphics[width=\linewidth]{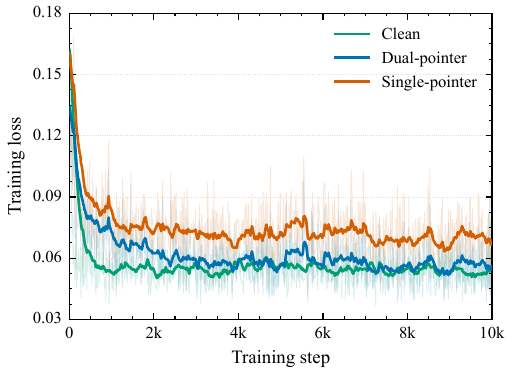}\par
        {\small\textbf{(b) Training loss}}
    \end{minipage}
    \begingroup
    \captionof{figure}{\textbf{Efficiency and optimization.}
    (a) Peak GPU memory and forward-pass time on one H100.
    (b) Dual-pointer loss approaches clean-context loss and remains below single-pointer loss.}
    \label{fig:history-scaling}
    \label{fig:training-loss-comparison}
    \endgroup
    \par\medskip
    \includegraphics[width=0.98\linewidth]{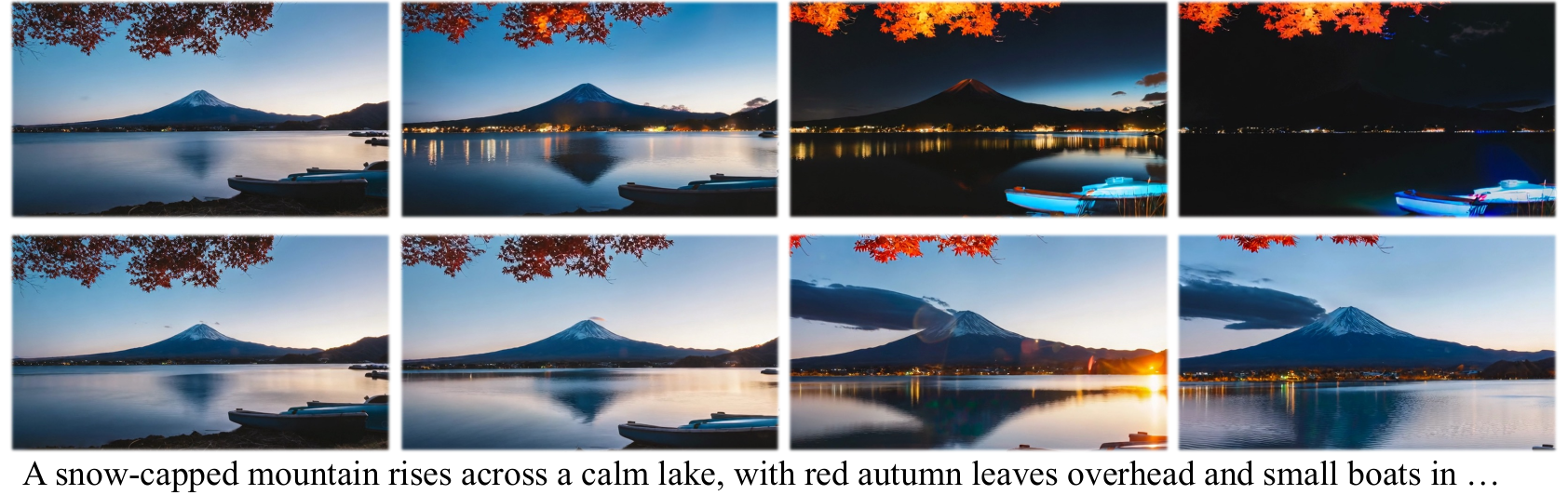}
    \begingroup
    \captionof{figure}{\textbf{Anti-drifting via dual-pointer denoising.}
    Generated frames from clean context (top) and dual-pointer denoising
    (bottom), with time progressing from left to right.}
    \label{fig:clean-context-dual-pointer}
    \endgroup
\end{table}

\subsection{Ablation Studies}
We examine history efficiency, granularity, 
and context denoising. Unless otherwise specified, 
variants use the same resolution of $640\times368$, a maximum history of 20 seconds, and 10K training steps. Evaluation uses fixed subsets of VBench and the off-screen subject reappearance benchmark.

\paragraph{Efficiency with longer history.}
We measure GPU memory usage and single-forward-pass time across history lengths on one H100 GPU with FlashAttention-3~\citep{fa3-2024}. In Fig.~\ref{fig:history-scaling}(a), a one-minute history requires memory comparable to the standard 480p, five-second short-video configuration and has a shorter forward time. This measured scaling supports end-to-end training and inference with extended historical context.

\paragraph{History granularity.}
Table~\ref{tab:history-granularity}(a) varies the history block shape and MLP bottleneck dimension. 
At the same nominal history-token budget, $(16,4,8,8)$ and $(16,16,4,4)$ produce 
similar overall VBench scores, with minor differences in memory metrics. Reducing the MLP bottleneck to 128 dimensions lowers VBench and character consistency, supporting the value of sufficient projection capacity.
Finer spatial grouping, from $(16,2,8,8)$ to $(16,2,4,4)$, improves face consistency when the subject is absent from the first frame. However, it requires approximately four times as many history tokens and does not improve the overall VBench score. We use $(16,2,8,8)$ as the default balance between efficiency and generation quality.

\paragraph{Clean context, single pointer, or dual pointer?}
We compare clean context, single-pointer denoising (adding noise to context tokens), and dual-pointer denoising using changes in VBench metrics between the first and last two-second windows. Dual-pointer denoising performs best on all four reported metrics (Tab.~\ref{tab:context-drift}(b)). Clean context also maintains stable subject and background consistency, suggesting that native long-video training already helps limit error accumulation.

The training curves in Fig.~\ref{fig:training-loss-comparison}(b) show that dual-pointer denoising and clean context reach similar final losses, both below single-pointer denoising. This is consistent with the paired-view shortcut making noisy context easier to use during optimization.

Figure~\ref{fig:clean-context-dual-pointer} illustrates the difference during rollout. Both variants exhibit color drift, but small shifts accumulate under clean context until the scene becomes excessively dark. With dual-pointer denoising, the drift is less pronounced in the final chunk, illustrating reduced error propagation across chunks.

\FloatBarrier
\section{Related Work}

\paragraph{Autoregressive video generation and temporal drift.}
Autoregressive video models support streaming~\citep{helios2026}, generative transmission~\citep{aiflow2025,aiflow2026}, and interactive applications~\citep{longlive2025,longlive2026}, but conditioning on generated history can propagate prediction errors. Prior work addresses this through more robust context conditioning, error-aware training, and planning. Noise perturbation uses scheduled noise levels for historical and newly generated frames to improve tolerance to imperfect context~\citep{df2024,hgvd2025}. Error-aware training instead simulates erroneous histories~\citep{helios2026} or feeds model predictions back as context~\citep{self-forcing2025,rolling-forcing2025}.

Planning methods generate farther-ahead frames or organize the rollout globally to limit the accumulation of local errors~\citep{tempomaster2026,framepack2025}. Recent error-blocking work uses the same noise strength for historical and newly generated frames to restrict error propagation~\citep{hiar2026}. Our dual-pointer design uses separate context and target noise schedules, together with paired context views that expose the denoising direction. The prediction objective remains focused on the current chunk.

\paragraph{Historical access and memory representation.}
Long-term memory must balance the cost of reading history with the information retained for future generation. Keyframe- and view-based methods retrieve frame-level observations using temporal or spatial cues, including FOV overlap, camera pose, and surfel associations~\citep{cam2025,vmem2025,worldmem2025}. Explicit 3D memories, such as point clouds and depth-guided latent caches, preserve scene structure across viewpoints~\citep{spatia2025,latentspatialmemory2026}. These approaches organize historical access around selected observations or geometric structure.

Other methods reduce the memory footprint through packed frame contexts, compact history embeddings, recurrent states, or learned query interfaces~\citep{framepack2025,tinyhistory2025,longcontextssm2025,car2025,memlearner2026}. Sparse attention reduces reading cost through chunk routing or token grouping~\citep{moc2026,moga2026}. \method packs local spatiotemporal blocks into high-dimensional tokens and exposes all blocks to native self-attention. The block representations are learned jointly with video generation, combining full-history coverage with a shorter conditioning sequence.
\section{Conclusion}

We introduced \method, which learns long-term video memory by packing local spatiotemporal history into high-dimensional tokens processed through native self-attention. This design reduces historical token count while preserving full-history coverage, allowing the video model to jointly learn how historical information is represented and used. Dual-pointer denoising complements this memory mechanism by assigning separate noise schedules to recent context and target frames to mitigate error propagation during autoregressive rollout. \method achieves the highest overall VBench-Long score of 80.1 among the evaluated methods and leads on all four consistency metrics for off-screen subject reappearance, while remaining competitive on scene revisitation. These results demonstrate that effective long-term memory can be learned from general-purpose video data.

Reliable memory use in out-of-domain settings remains an open challenge. Future work will explore more demanding revisitation scenarios, diverse downstream applications, and learning from in-context demonstrations.

\bibliographystyle{plainnat}
\bibliography{main}

\clearpage
\appendix
\section*{Appendix}
\section{Camera Control}
\label{app:camera-control}

For the static-scene revisitation evaluation, we adapt \method using
36k video clips from SpatialVID-HQ\allowbreak~\citep{spatialvid2026}, paired with
text captions and camera poses. 

\paragraph{Pose preprocessing.}
SpatialVID-HQ provides sparse camera poses as quaternions and translations 
in world-to-camera coordinates. We convert them to camera-to-world 
coordinates, interpolate the poses to the sampled frames, and 
associate one camera pose with each temporal latent using the last 
source frame in the corresponding temporal group 
(with a VAE temporal stride of 4, we sample every 4th frame).
During both training and inference, camera-to-world poses are expressed relative to the first 
frame of the current chunk. For a window of $F$ temporal latents, 
the camera input is represented as an $F \times 12$ array containing the 
first three rows of each relative camera-to-world matrix in row-major 
order.

\paragraph{Fused-PRoPE.}
We implement camera conditioning through fused-PRoPE~\citep{mosaicmem2026}
within the existing self-attention operation. 
We use fixed normalized focal lengths
$\bar f_x = f_x/W = 0.50505$ and
$\bar f_y = f_y/H = 0.93795$ at
$W\times H = 832\times448$. 
Let the diagonal matrix $\mathbf{K}=\operatorname{diag}(\bar f_x,\bar f_y,1,1)$ 
denote the projection transform.
Let $\mathbf{C}_i$ denote
the relative camera-to-world matrix for the frame containing token $i$.
We form the projection and its inverse as
\begin{equation}
    \mathbf{P}_i = \mathbf{K}\mathbf{C}_i^{-1},
    \qquad
    \mathbf{P}_i^{-1} = \mathbf{C}_i\mathbf{K}^{-1}.
\end{equation}
The same per-frame matrices are shared by all spatial patch tokens
in that frame. Each $4\times4$ transform is applied independently
to consecutive groups of four feature channels. Writing this repeated 
transform as $\mathcal{P}_i = I_{d/4}\otimes\mathbf{P}_i$ for feature width $d$,
we first apply the standard 3D RoPE to the queries and keys
to preserve the pretrained DiT's spatiotemporal modeling capabilities.
We then apply PRoPE directly to the resulting queries and keys,
denoted by $q_i$ and $k_i$, together with the values $v_i$:
\begin{equation}
    \widetilde{q}_i = \mathcal{P}_i^{\mathsf T}q_i,
    \qquad
    \widetilde{k}_i = \mathcal{P}_i^{-1}k_i,
    \qquad
    \widetilde{v}_i = \mathcal{P}_i^{-1}v_i.
\end{equation}
We then perform a single self-attention operation using
$\widetilde{q}_i$, $\widetilde{k}_i$, and $\widetilde{v}_i$.
For its output $\widetilde{o}_i$, we apply
$o_i = \mathcal{P}_i\widetilde{o}_i$ before the existing output projection.

Note that the camera-training clips have a median duration of 12.35 
seconds and a maximum duration of 15.04 seconds, both shorter than 
those in the preceding training stages. Nevertheless, we find that the 
camera-control capability generalizes to 30-second generation, thanks to 
the prior provided by PRoPE.

\section{Training Data}
\label{app:training-data}

\begin{figure}[htbp]
    \centering
    \includegraphics[width=\linewidth]{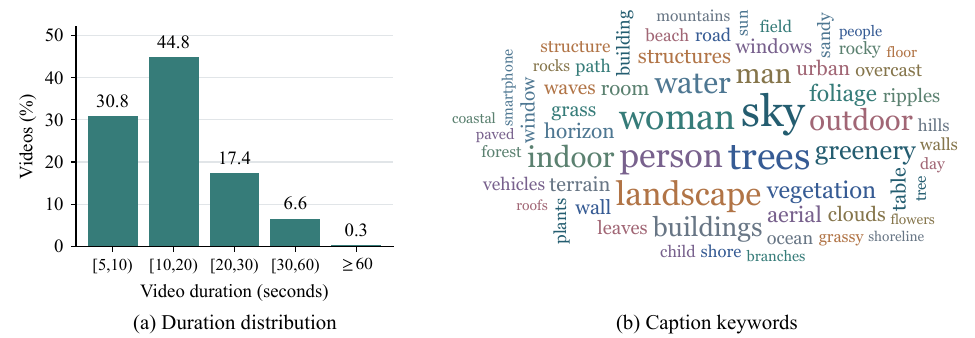}
    \caption{Overview of the general-purpose video corpus.
    (a) Percentage of videos in each duration interval.
    (b) Word cloud of the 60 most frequent words in the captions after filtering out template terms such as ``captures'' and ``environment''.
}
    \label{fig:training-data-overview}
\end{figure}

\paragraph{General-purpose video corpus.}
We train \method on an internal corpus of 2 million distinct videos,
totaling approximately 8000 hours. The corpus consists primarily of
web-sourced footage from commercial stock-video collections (e.g., iStock),
supplemented by other supplied datasets, public stock-video websites
(e.g., Pexels and Mixkit), and publicly available video datasets
(e.g., Koala-36M \citep{koala2025}).
Videos have a median duration of 13.15 seconds, and nearly all last
between 5 and 60 seconds. Fig.~\ref{fig:training-data-overview}(a) summarizes
the duration distribution.
We further provide a qualitative overview of the training video content by visualizing the 
distribution of caption keywords as a word cloud in Fig.~\ref{fig:training-data-overview}(b).

\paragraph{Data preparation and sampling.}
We filter the source videos using several quality signals, including 
aesthetics, motion magnitude, sharpness, image quality, and text overlays. 
The videos are resampled to 16 fps, resized while preserving their aspect 
ratio, and center-cropped to $832 \times 448$ pixels. We then use 
Qwen3-VL-30B-A3B~\citep{qwen3vl2025} to generate dense captions for temporal segments of 
five seconds each.
Training examples are sampled at the caption-segment level. For each 
example, the target generation window lies within the selected caption 
segment, while the visual history is drawn from the immediately preceding 
portion of the same video. Although this history may span multiple 
earlier caption segments, only the caption of the target segment is provided 
to the model, which also aligns with our attention mechanism.
Specifically, to improve training stability, we mix in image-to-video 
samples at a 10\% ratio, which corresponds to the case of zero history length.

\paragraph{No use of any memory supervision.}
We do not use manually constructed memory labels, such as annotations
identifying scene revisits through field-of-view overlap or specifying
which past frames to retrieve. Historical observations serve only as
conditioning inputs, and the flow-matching loss is computed only on the
current generation chunk. The model therefore learns to represent and
use its history through the video denoising objective, without an
additional loss supervising memory retrieval.

\section{Evaluation Details}
\label{app:evaluation-details}

For VBench-Long, we use the official evaluation protocol without
modification~\citep{vbench++2024}. Below, we detail the protocols for
static-scene revisitation and off-screen subject reappearance.

\subsection{Static-Scene Revisitation}

\paragraph{Inputs and camera trajectory.}
We use the 100 reference images and their accompanying text prompts
from R2M-Bench~\citep{r2mbench2026}. 
In our experiments, we found that the original trajectories in R2M-Bench
are overly complex for our method and some baselines (e.g., Sana-WM and
LingBot-World-Fast), sometimes leading to degraded generation quality
and camera trajectory adherence. To fairly evaluate memory capability,
we construct a 30-second trajectory that rotates left by $90^\circ$
over the first 7.5 seconds, right by $180^\circ$ over the next 15 seconds,
and left by $90^\circ$ over the final 7.5 seconds, finally returning
to the initial orientation.
Before scoring, videos from all methods are resized while preserving
their aspect ratio and center-cropped to $832\times480$ pixels, with
their frame counts unchanged. This common evaluation resolution
allows pixel-level metrics to be computed under the same spatial
preprocessing.

\paragraph{Frame-pair construction.}
We evaluate three sets of frame pairs within each generated video.
Revisit pairs connect temporally separated frames with matching
camera orientations and positions.
Gap-matched frame pairs are sampled to match the temporal separations
of revisit pairs and serve as a baseline for measuring frame similarity
at comparable temporal gaps.
Short-range pairs measure local consistency, with temporal gaps
between 2 and $\max(6,0.02T)$ frames for a video of $T$ frames.
Following the original setting, we randomly sample 100 pairs per set. 

\paragraph{Evaluation metrics.}
We report eight metrics from R2M-Bench: PSNR, SSIM, LPIPS, DINO similarity, BoQ similarity, 
MVPR similarity, object appearance persistence, and object semantic persistence. 
The first six metrics compare frames at the image level. The last two 
metrics characterize object persistence: object appearance persistence 
measures the visual similarity between matched objects, while object 
semantic persistence measures how strongly the matched detections support 
the semantic concepts of the anchor objects.

For each frame pair, we detect anchor objects in the earlier frame using 
GroundingDINO~\citep{groundingdino2024} with a fixed vocabulary of object concepts, using box and 
text thresholds of $0.25$ and $0.20$, respectively, and a non-maximum 
suppression threshold of $0.6$. We retain at most $20$ anchors and 
discard objects whose area is less than $0.2\%$ of the image. SAM2 
provides a mask for each object, from which we extract a DINOv2~\citep{dinov2-2023} descriptor. 
In the later frame, we independently detect up to $5$ candidates for each 
anchor concept and compute their masks and descriptors.

For an anchor $u$ and a candidate $c$, we compute an appearance score for 
object appearance persistence:
\begin{equation}
    a_{u,c} = \operatorname{clamp}_{[0,1]}
    \left(\frac{1+\cos(d_u,d_c)}{2}\right),
\end{equation}
where $d_u$ and $d_c$ are their DINOv2 descriptors.
For object semantic persistence, the semantic score combines the candidate's GroundingDINO detection confidence $b_c$ with CLIP similarity~\citep{clip2021} between its image feature $g_c$ and the text feature $t_{\ell_u}$ of the anchor's concept $\ell_u$:
\begin{equation}
    s_{u,c} = \tfrac{1}{2}\operatorname{clamp}_{[0,1]}(b_c)
    + \tfrac{1}{2}\operatorname{clamp}_{[0,1]}
    \left(\frac{1+\cos(g_c,t_{\ell_u})}{2}\right).
\end{equation}
For each anchor, we select the candidate maximizing $0.7a_{u,c}+0.3s_{u,c}$ and accept the match only if this combined score is at least $0.5$.

Let $N$ be the total number of anchor objects, $\mathcal{M}$ the set 
of anchors with accepted matches, and $c_u^*$ the selected candidate for 
a matched anchor. The two persistence scores are
\begin{equation}
    \mathrm{AppPersist} = \frac{1}{N}
        \sum_{u\in\mathcal{M}} a_{u,c_u^*},
    \qquad
    \mathrm{SemPersist} = \frac{1}{N}
        \sum_{u\in\mathcal{M}} s_{u,c_u^*}.
\end{equation}
If no anchor is detected ($N=0$), both persistence scores are set to zero, 
and the frame pair is retained in the evaluation.

\paragraph{Memory gain and normalization.}
For a given metric and video $v \in \mathcal{V}$, let $R_v$,
$B_v$, and $S_v$ denote its mean scores over revisit, gap-matched,
and short-range pairs, respectively. We define
\begin{equation}
    \mathrm{MG}_v = i(R_v-B_v),
    \qquad
    \mathrm{DR}_v = i(S_v-B_v),
\end{equation}
where $i=1$ for higher-is-better measures and $i=-1$ for LPIPS.
MemoryGain (MG) measures the improvement on revisit pairs over the
gap-matched baseline. The dynamic range (DR) measures how much local
consistency exceeds this baseline. The Normalized Memory Ratio (NMR)
is then computed on the average of MGs and DRs:
\begin{equation}
    \mathrm{NMR} =
    \frac{\operatorname{mean}_{v\in\mathcal{V}}
          \mathrm{MG}_v}
         {\operatorname{mean}_{v\in\mathcal{V}}
          \mathrm{DR}_v+\epsilon}.
\end{equation}
 We report MG and NMR
separately for each of the eight measures.

\subsection{Off-Screen Subject Reappearance}
\label{app:offscreen-subject-reappearance}

\paragraph{Benchmark construction.}
We collect subject and background reference images from Pexels and generate 
candidate histories with MiniMax-H3-Ref2V \citep{minimaxh3_2026}. We then 
manually filter the generated histories to obtain $64$ high-quality cases 
covering $40$ subjects and $10$ backgrounds.

\paragraph{The history for evaluation.}
We use $50$ sampling steps at a native resolution of $960\times544$ and resize 
the outputs to $1280\times720$. We generate $15$-second videos to serve as 
historical context for the baselines. We present two prompt templates: one for 
cases where the subject appears in the first frame and one for cases where the 
subject does not appear in the first frame
(Fig.~\ref{fig:history-generation-prompt}).

\begin{figure}[p]
    \centering
    \begingroup
    \setlength{\fboxsep}{6pt}
    \setlength{\fboxrule}{0.4pt}
    \fontsize{8}{9.0}\selectfont\ttfamily\raggedright
    \fbox{\begin{minipage}{\dimexpr\linewidth-2\fboxsep-2\fboxrule\relax}
    \setlength{\parindent}{0pt}
    \setlength{\parskip}{1pt}
    {\normalfont\small\bfseries (a) Initially present}\par\smallskip
At the beginning of the target video, <Picture 1> is fully referenced.

subject\_definitions:\par
<Subject 1> is a South Asian woman in a yellow embroidered dress, whose identity, face, hair, body
proportions, and clothing come from <Picture 1>.
<Picture 1> is the complete visual reference for <Subject 1> and the scene. ...

summary:\par
[reference generation] ... The person is already present at the start, performs one simple upper-body
action, then exits laterally while the fixed camera holds on the same unchanged environment ...

retention\_analysis:\par
<Subject 1> (visible throughout [Shot 1]): fully\_preserved - preserve the referenced face, hairstyle,
clothing, body proportions, and identity while allowing new motion.

detailed\_description:\par
[Shot 1] At 0.00 seconds, <Subject 1> is already present and facing the camera. ...
\par
From 0.00 to 1.00 seconds, the person smiles naturally and gives one clear, decisive one-handed wave.
\par
From 1.00 to 1.50 seconds, the person immediately settles and turns naturally into a slight
three-quarter view to begin leaving.
\par
From 1.50 to 5.00 seconds, the person takes decisive, continuous lateral steps toward screen right
and leaves the frame ...
\par
By 5.00 seconds the person is completely out of frame; from 5.00 to 15.00 seconds the fixed camera
holds on only the unchanged empty scene and background layout from <Picture 1>.
\par
...
    \end{minipage}}

    \par\vspace{8pt}
    \fbox{\begin{minipage}{\dimexpr\linewidth-2\fboxsep-2\fboxrule\relax}
    \setlength{\parindent}{0pt}
    \setlength{\parskip}{1pt}
    {\normalfont\small\bfseries (b) Initially absent}\par\smallskip
At the beginning of the target video, the visible environment, lighting, colors, and camera
composition from <Picture 1> are fully referenced; <Subject 1> is outside the frame and the first
frame is an empty scene.

subject\_definitions:\par
<Subject 1> is a Black man with short hair in a white T-shirt and blue trousers, whose identity,
face, hair, body proportions, and clothing come from <Picture 1>.
<Picture 1> is the complete visual reference for <Subject 1> and the scene. Preserve ...

summary:\par
[reference generation] The target is one continuous 15-second, 16:9 horizontal video using <Picture 1>
as the complete subject-and-scene reference. The video starts with the same scene empty, with
<Subject 1> outside the frame; the person enters from off-screen, faces the camera, smiles, then walks
naturally out of the scene while the fixed camera holds on the same unchanged environment ...

detailed\_description:\par
[Shot 1] At 0.00 seconds, the frame shows the empty scene from <Picture 1> and <Subject 1> is
completely outside the frame. ...
\par
From 0.00 to 1.50 seconds, show only the empty reference scene; <Subject 1> remains outside.
\par
From 1.50 to 3.00 seconds, <Subject 1> briskly enters from outside the screen-right edge ...
\par
From 3.00 to 3.60 seconds, the person quickly settles and faces the camera with an unobstructed face.
\par
From 3.60 to 5.00 seconds, the person faces the camera, smiles naturally, and keeps both hands relaxed.
\par
From 5.00 to 6.60 seconds, the person naturally turns and walks briskly toward screen left, leaving
the scene completely ...
\par
From 6.60 to 15.00 seconds, the fixed camera holds only the same unchanged empty reference scene.
\par
... 
    \end{minipage}}
    \endgroup
    \caption{Example history-generation prompts with the subject
    (a) initially present and (b) initially absent. Ellipses indicate omitted text.}
    \label{fig:history-generation-prompt}
\end{figure}

Note that we supply the same source histories to all methods, but the amount of 
history accessible through their continuation interfaces differs. Specifically, 
\method and the two LingBot-World variants access the full 15-second history. 
LongLive 2.0 retains the first 29 pixel frames and the last 96 pixel frames, 
while FramePack-F1 and Helios retain the initial frame and the most recent 73 
pixel frames. 

\paragraph{Evaluation metrics.}
We compare the generated continuation with the reference history using four 
metrics from MSVBench~\citep{msvbench2026}. We first apply GroundingDINO~\citep{groundingdino2024} and
SAM~\citep{sam2023} to detect people in each frame and group consecutive detections into 
intervals of subject visibility. 

For face and character consistency, we select one interval of subject
visibility from the reference history and one from the generated
continuation, then uniformly sample 10 frames from each interval.
We compare every sampled history frame with every sampled continuation
frame, yielding $10\times10=100$ pairs without temporal alignment.

For face consistency, we detect faces using DeepFace's OpenCV detector
and extract VGG-Face embeddings. Given a history embedding $f_i^H$
and a continuation embedding $f_j^G$, we compute their cosine distance
and average over all pairs:
\begin{equation}
    d_{ij}^{\mathrm{face}}
    = 1-\frac{(f_i^H)^{\mathsf T}f_j^G}
    {\lVert f_i^H\rVert_2\lVert f_j^G\rVert_2},
    \qquad
    D_{\mathrm{face}}
    = \frac{1}{100}\sum_{i=1}^{10}\sum_{j=1}^{10}
    d_{ij}^{\mathrm{face}}.
\end{equation}
For character consistency, we crop the largest detected person in
each sampled frame and extract a whole-person appearance embedding
using InceptionNeXt~\citep{inceptionnext2024}. Given history and continuation embeddings
$z_i^H$ and $z_j^G$, we compute their Euclidean distance and average
over the same frame pairs:
\begin{equation}
    d_{ij}^{\mathrm{char}} = \lVert z_i^H-z_j^G\rVert_2,
    \qquad
    D_{\mathrm{char}}
    = \frac{1}{100}\sum_{i=1}^{10}\sum_{j=1}^{10}
    d_{ij}^{\mathrm{char}}.
\end{equation}

For background consistency, we select one person-free frame from
each video. From the history, we take the frame at the midpoint
of the empty interval following the selected interval of subject
visibility. From the continuation, we take the earliest available
empty frame before the subject reappears. We compute and report the DreamSim~\citep{dreamsim2023}
perceptual distance between these two frames.

For clothes consistency, we select one frame from each video and
crop the largest detected person in each frame. We provide both
crops, together with the subject's textual description, to
Gemini-3.1-pro-preview. The judge first verifies that the subject
is present in both images, then rates the consistency of clothing
and colors on a scale from 1 to 5, accounting for changes in
viewpoint and lighting. Higher scores indicate greater consistency.
Note that cases in which the subject is not generated in the continuation are
excluded from the affected metrics, since the corresponding measurements
cannot be defined.

\section{On the context noise schedule}
\label{app:context-noise-schedule}

\begin{figure}[htbp]
    \centering
    \includegraphics[width=\linewidth]{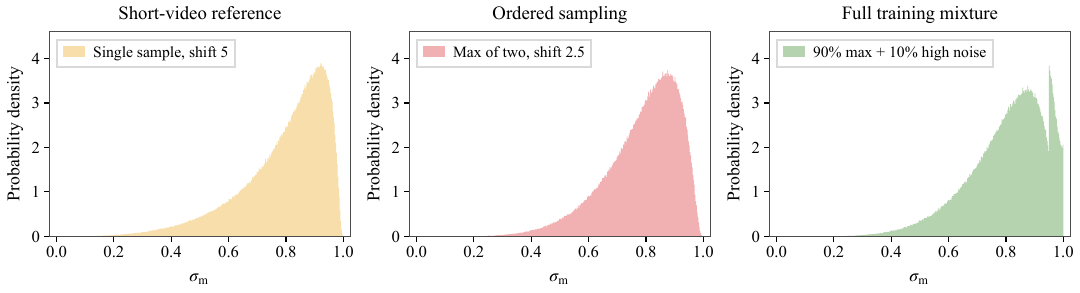}
    \caption{Probability densities of the training noise level.
    The reference uses a single shifted logit-normal sample with shift $5$.
    Our min/max branch takes the larger of two samples with shift $2.5$;
    the full training mixture additionally includes 10\% uniform samples
    from $[0.95,1]$.}
    \label{fig:context-noise-density}
\end{figure}

During training, we sample the context and current-chunk noise levels jointly.
Let $r_s(u)=\frac{su}{1+(s-1)u}$ denote the shift transform. We draw
\begin{equation}
    z_1,z_2 \overset{\mathrm{i.i.d.}}{\sim} \mathcal{N}(0,1),
    \qquad
    S_i=r_{s_{\mathrm{train}}}\!\left(\operatorname{sigmoid}(z_i)\right),
    \qquad s_{\mathrm{train}}=2.5.
\end{equation}
The two noise levels are then assigned as
\begin{equation}
    (\sigma_{\mathrm c},\sigma_{\mathrm m})=
    \begin{cases}
        \bigl(\min(S_1,S_2),\max(S_1,S_2)\bigr),
            & \text{with probability }0.9,\\
        (H,H),\quad H\sim\mathcal{U}(0.95,1),
            & \text{with probability }0.1.
    \end{cases}
\end{equation}
Since the original logit-normal distribution assigns relatively little 
probability mass to the high-noise region, 
we introduce the second branch to sample additional high-noise training examples.
However, taking the maximum of two samples shifts the current-chunk distribution
toward higher noise levels. 
To compensate for this effect, we set $s_{\mathrm{train}}=2.5$, thereby keeping 
its distribution close to that of the single-sample reference under the 
commonly used shift-5 configuration for five-second clips at $832\times448$ 
resolution.
Fig.~\ref{fig:context-noise-density} compares their probability densities,
showing both the min/max branch and the full training
mixture. 

\begin{figure}[!htbp]
    \centering
    \begin{minipage}[t]{0.53\linewidth}
        \centering
        \vspace{0pt}
        \includegraphics[width=\linewidth]{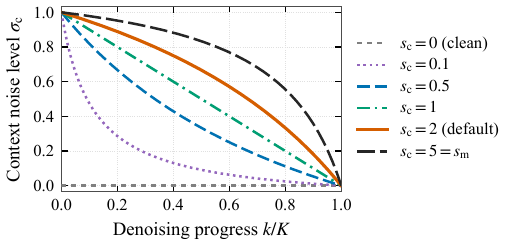}
        \caption{Context noise schedules with $s_{\mathrm m}=5$.
        Larger $s_{\mathrm c}$ keeps the context noisier;
        $s_{\mathrm c}=0$ is clean and $s_{\mathrm c}=5$ matches
        the current-chunk schedule.}
        \label{fig:context-noise-schedule}
    \end{minipage}\hfill
    \begin{minipage}[t]{0.46\linewidth}
        \centering
        \vspace{0pt}
        \captionof{table}{Temporal drift with $s_{\mathrm m}=5$.
        $\Delta=\mathrm{last\ 2\,s}-\mathrm{first\ 2\,s}$, averaged
        over videos; negative values indicate degradation.
        Bold marks the best result per column.}
        \label{tab:context-shift-drift}
        \smallskip
        \begingroup
    \fontsize{8}{9}\selectfont
    \setlength{\tabcolsep}{1pt}
    \renewcommand{\arraystretch}{1.1}
    \begin{tabular}{@{}lrrrr@{}}
        \toprule
        $s_{\mathrm c}$
        & \shortstack{\textbf{Subject}\\$\Delta\uparrow$}
        & \shortstack{\textbf{Background}\\$\Delta\uparrow$}
        & \shortstack{\textbf{Imaging}\\$\Delta\uparrow$}
        & \shortstack{\textbf{Aesthetic}\\$\Delta\uparrow$} \\
        \midrule
        0 & $-$0.0038 & $-$0.0012 & $-$0.0346 & $-$0.0554 \\
        0.1 & $+$0.0001 & $-$0.0002 & $-$0.0121 & $-$0.0565 \\
        0.5 & $+$\textbf{0.0019} & $+$\textbf{0.0024} & $-$0.0214 & $-$0.0445 \\
        1 & $-$0.0017 & $-$0.0002 & $-$\textbf{0.0011} & $-$0.0491 \\
        2 & $+$0.0011 & 0.0000 & $-$0.0012 & $-$0.0335 \\
        5 & $+$0.0014 & $+$0.0020 & $-$0.0012 & $-$\textbf{0.0262} \\
        \bottomrule
    \end{tabular}
\endgroup

    \end{minipage}
\end{figure}

We further evaluate how the context noise shift affects generation quality at 
inference time. Fixing the current-chunk shift at $s_{\mathrm m}=5$, we compare 
context shifts $s_{\mathrm c}\in\{0,0.1,0.5,1,2,5\}$. Here, $s_{\mathrm c}=0$
denotes a fully clean context throughout denoising, while $s_{\mathrm c}=5$ 
keeps the context noise level identical to that of the current chunk at every 
denoising step. Fig.~\ref{fig:context-noise-schedule} illustrates the
corresponding schedules with a normalized denoising progress.

\begin{table}[!htbp]
    \centering
    \caption{\textbf{Effect of the context noise shift on VBench scores.}
    The current-chunk shift is fixed at $s_{\mathrm m}=5$. Bold marks the best result per column.}
    \label{tab:context-shift-vbench}
    \smallskip
    \small
    \setlength{\tabcolsep}{3pt}
    \renewcommand{\arraystretch}{1.1}
    \begin{tabular}{@{}lrrrrrrr@{}}
        \toprule
        \shortstack[l]{\textbf{Context shift}\\$s_{\mathrm c}$}
        & \shortstack{\textbf{Total}\\\textbf{Score}}
        & \shortstack{\textbf{Subject}\\\textbf{Consistency}}
        & \shortstack{\textbf{Background}\\\textbf{Consistency}}
        & \shortstack{\textbf{Motion}\\\textbf{Smoothness}}
        & \shortstack{\textbf{Dynamic}\\\textbf{Degree}}
        & \shortstack{\textbf{Imaging}\\\textbf{Quality}}
        & \shortstack{\textbf{Aesthetic}\\\textbf{Quality}} \\
        \midrule
        0 & 79.59 & 97.40 & 96.81 & 98.02 & \textbf{47.73} & 69.50 & 57.69 \\
        0.1 & 79.78 & 97.58 & 96.92 & 98.11 & 46.88 & 70.07 & 57.92 \\
        0.5 & 79.91 & 97.72 & 97.04 & 98.22 & 46.31 & 69.88 & 58.43 \\
        1 & \textbf{80.01} & 97.58 & 96.99 & 98.19 & 46.31 & \textbf{70.77} & 58.37 \\
        2 & 79.97 & 97.73 & 97.01 & 98.22 & 44.89 & 70.68 & 58.65 \\
        5 & 79.89 & \textbf{97.79} & \textbf{97.05} & \textbf{98.32} & 42.33 & 70.71 & \textbf{59.03} \\
        \bottomrule
    \end{tabular}
\end{table}

Tabs.~\ref{tab:context-shift-drift} and~\ref{tab:context-shift-vbench} report 
temporal drift and overall VBench scores, respectively. 
We observe that small context noise shifts (0 and 0.1) expose a cleaner context 
too early, which introduces prediction errors into more denoising steps and 
degrades the temporal drift metric. This is consistent with our design hypothesis. 
As the context noise shift increases, the generated videos become increasingly 
static, as reflected by a lower dynamic degree (from 47.73 to 42.33). 
This finding echoes \citet{hiar2026}: a noisier history may make it harder 
for the model to determine the motion structure of newly generated frames in 
early denoising steps, leading to static videos. A moderate shift, such as 1 or 2, 
achieves the best trade-off between these two extremes, yielding the highest 
VBench scores and the most favorable temporal drift metrics. 
We therefore adopt $s_{\mathrm c}=2$ as the default setting in our experiments.

\section{Additional Qualitative Results}
\label{app:qualitative-results}

% Break qualitative comparisons only between complete method rows.  The
% source PDFs stay unchanged; trim/clip preserves their native vector labels.
\begingroup
\renewcommand{\LTcaptype}{}
\renewcommand{\arraystretch}{0}
\setlength{\LTpre}{4pt}
\setlength{\LTpost}{7pt}
\setlength{\LTleft}{0pt}
\setlength{\LTright}{0pt}
% Reserve the figure number before its cases; typeset the caption after
% the final image row without incrementing the counter a second time.
\makeatletter
\newcommand{\appfigurecaption}[1]{%
    \@makecaption{\fnum@figure}{#1}%
}
\newcommand{\appcaselabel}[2]{%
    \protected@edef\@currentlabel{\thefigure}%
    \def\@currentlabelname{#1}%
    \label{#2}%
}
\makeatother
\newenvironment{appresultcase}[2]{%
% Give unnumbered longtables distinct width caches and hyperlink targets.
    \stepcounter{LT@tables}%
    \begin{longtable}{@{}p{\linewidth}@{}}
    {\small\bfseries #1}\appcaselabel{#1}{#2}\\[3pt]
    \endfirsthead
    {\small\bfseries Figure \thefigure: #1 (continued)}\\[3pt]
    \endhead
    \multicolumn{1}{@{}r@{}}{\footnotesize Continued on the next page.}\\
    \endfoot
    \endlastfoot
}{%
    \end{longtable}
}

We provide additional examples of off-screen subject reappearance,
static-scene revisitation, and image-to-video generation.

\subsection{Off-Screen Subject Reappearance}

\refstepcounter{figure}
\label{fig:app-subject-white}

\begin{appresultcase}{Case 1: White top}{fig:app-subject-white-case}
    \includegraphics[page=1,trim=0 996.290 0 10.186,clip,width=\linewidth]{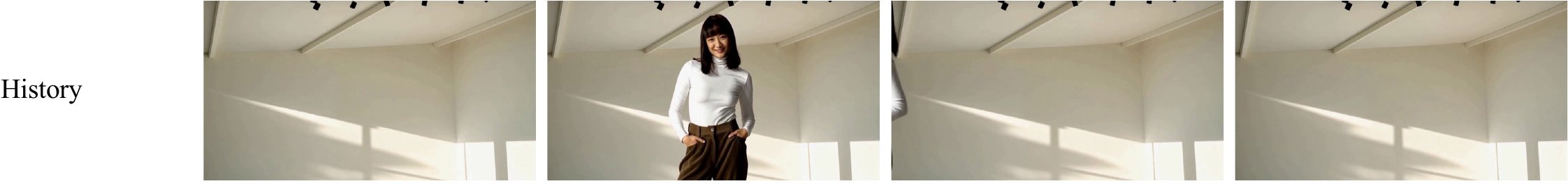}\\*
    \includegraphics[page=1,trim=0 865.267 0 137.710,clip,width=\linewidth]{assets/subject_revisit.pdf}\\
    \includegraphics[page=1,trim=0 734.243 0 268.733,clip,width=\linewidth]{assets/subject_revisit.pdf}\\
    \includegraphics[page=1,trim=0 603.220 0 399.757,clip,width=\linewidth]{assets/subject_revisit.pdf}\\
    \includegraphics[page=1,trim=0 472.196 0 530.780,clip,width=\linewidth]{assets/subject_revisit.pdf}\\
    \includegraphics[page=1,trim=0 341.172 0 661.804,clip,width=\linewidth]{assets/subject_revisit.pdf}\\
    \includegraphics[page=1,trim=0 210.149 0 792.828,clip,width=\linewidth]{assets/subject_revisit.pdf}\\*
    \includegraphics[page=1,trim=0 27.800 0 923.851,clip,width=\linewidth]{assets/subject_revisit.pdf}\\
\end{appresultcase}

\begin{appresultcase}{Case 2: Tan sweater and eyeglasses}{fig:app-subject-glasses}
    \includegraphics[page=2,trim=0 996.290 0 10.186,clip,width=\linewidth]{assets/subject_revisit.pdf}\\*
    \includegraphics[page=2,trim=0 865.267 0 137.710,clip,width=\linewidth]{assets/subject_revisit.pdf}\\
    \includegraphics[page=2,trim=0 734.243 0 268.733,clip,width=\linewidth]{assets/subject_revisit.pdf}\\
    \includegraphics[page=2,trim=0 603.220 0 399.757,clip,width=\linewidth]{assets/subject_revisit.pdf}\\
    \includegraphics[page=2,trim=0 472.196 0 530.780,clip,width=\linewidth]{assets/subject_revisit.pdf}\\
    \includegraphics[page=2,trim=0 341.172 0 661.804,clip,width=\linewidth]{assets/subject_revisit.pdf}\\
    \includegraphics[page=2,trim=0 210.149 0 792.828,clip,width=\linewidth]{assets/subject_revisit.pdf}\\*
    \includegraphics[page=2,trim=0 27.800 0 923.851,clip,width=\linewidth]{assets/subject_revisit.pdf}\\
\end{appresultcase}

\begin{appresultcase}{Case 3: Formal suit}{fig:app-subject-suit}
    \includegraphics[page=3,trim=0 996.290 0 10.186,clip,width=\linewidth]{assets/subject_revisit.pdf}\\*
    \includegraphics[page=3,trim=0 865.267 0 137.710,clip,width=\linewidth]{assets/subject_revisit.pdf}\\
    \includegraphics[page=3,trim=0 734.243 0 268.733,clip,width=\linewidth]{assets/subject_revisit.pdf}\\
    \includegraphics[page=3,trim=0 603.220 0 399.757,clip,width=\linewidth]{assets/subject_revisit.pdf}\\
    \includegraphics[page=3,trim=0 472.196 0 530.780,clip,width=\linewidth]{assets/subject_revisit.pdf}\\
    \includegraphics[page=3,trim=0 341.172 0 661.804,clip,width=\linewidth]{assets/subject_revisit.pdf}\\
    \includegraphics[page=3,trim=0 210.149 0 792.828,clip,width=\linewidth]{assets/subject_revisit.pdf}\\*
    \includegraphics[page=3,trim=0 50.840 0 923.851,clip,width=\linewidth]{assets/subject_revisit.pdf}\\
\end{appresultcase}

\begin{appresultcase}{Case 4: Yellow dress, piano}{fig:app-subject-yellow-one}
    \includegraphics[page=4,trim=0 996.290 0 10.186,clip,width=\linewidth]{assets/subject_revisit.pdf}\\*
    \includegraphics[page=4,trim=0 865.267 0 137.710,clip,width=\linewidth]{assets/subject_revisit.pdf}\\
    \includegraphics[page=4,trim=0 734.243 0 268.733,clip,width=\linewidth]{assets/subject_revisit.pdf}\\
    \includegraphics[page=4,trim=0 603.220 0 399.757,clip,width=\linewidth]{assets/subject_revisit.pdf}\\
    \includegraphics[page=4,trim=0 472.196 0 530.780,clip,width=\linewidth]{assets/subject_revisit.pdf}\\
    \includegraphics[page=4,trim=0 341.172 0 661.804,clip,width=\linewidth]{assets/subject_revisit.pdf}\\
    \includegraphics[page=4,trim=0 210.149 0 792.828,clip,width=\linewidth]{assets/subject_revisit.pdf}\\*
    \includegraphics[page=4,trim=0 27.800 0 923.851,clip,width=\linewidth]{assets/subject_revisit.pdf}\\
\end{appresultcase}

\begin{appresultcase}{Case 5: Yellow dress, dark studio}{fig:app-subject-yellow-two}
    \includegraphics[page=5,trim=0 996.290 0 10.186,clip,width=\linewidth]{assets/subject_revisit.pdf}\\*
    \includegraphics[page=5,trim=0 865.267 0 137.710,clip,width=\linewidth]{assets/subject_revisit.pdf}\\
    \includegraphics[page=5,trim=0 734.243 0 268.733,clip,width=\linewidth]{assets/subject_revisit.pdf}\\
    \includegraphics[page=5,trim=0 603.220 0 399.757,clip,width=\linewidth]{assets/subject_revisit.pdf}\\
    \includegraphics[page=5,trim=0 472.196 0 530.780,clip,width=\linewidth]{assets/subject_revisit.pdf}\\
    \includegraphics[page=5,trim=0 341.172 0 661.804,clip,width=\linewidth]{assets/subject_revisit.pdf}\\
    \includegraphics[page=5,trim=0 210.149 0 792.828,clip,width=\linewidth]{assets/subject_revisit.pdf}\\*
    \includegraphics[page=5,trim=0 27.800 0 923.851,clip,width=\linewidth]{assets/subject_revisit.pdf}\\
\end{appresultcase}

\begin{appresultcase}{Case 6: Burgundy sweater}{fig:app-subject-burgundy}
    \includegraphics[page=6,trim=0 996.290 0 10.186,clip,width=\linewidth]{assets/subject_revisit.pdf}\\*
    \includegraphics[page=6,trim=0 865.267 0 137.710,clip,width=\linewidth]{assets/subject_revisit.pdf}\\
    \includegraphics[page=6,trim=0 734.243 0 268.733,clip,width=\linewidth]{assets/subject_revisit.pdf}\\
    \includegraphics[page=6,trim=0 603.220 0 399.757,clip,width=\linewidth]{assets/subject_revisit.pdf}\\
    \includegraphics[page=6,trim=0 472.196 0 530.780,clip,width=\linewidth]{assets/subject_revisit.pdf}\\
    \includegraphics[page=6,trim=0 341.172 0 661.804,clip,width=\linewidth]{assets/subject_revisit.pdf}\\
    \includegraphics[page=6,trim=0 210.149 0 792.828,clip,width=\linewidth]{assets/subject_revisit.pdf}\\*
    \includegraphics[page=6,trim=0 50.840 0 923.851,clip,width=\linewidth]{assets/subject_revisit.pdf}\\*
    \begin{minipage}{\linewidth}
    \appfigurecaption{Off-screen subject reappearance across six cases.
Each case shows the shared history above \method and all six baselines.
Zoom in to compare the returning subject's face, clothing and motion with the history.}
    \end{minipage}\\
\end{appresultcase}

\subsection{Static-Scene Revisitation}
\label{app:qualitative-revisit}

Fig.~\ref{fig:app-revisit-lakeside} shows \method and all four
camera-controlled baselines across three scenes.
For every method, the four columns use frames at
$5$, $10$, $20$, and $25$ seconds, respectively, from its generated video.
Under the prescribed trajectory in Appendix~\ref{app:evaluation-details},
the $5/10$-second pair shares a target yaw of $-60^\circ$, while the
$20/25$-second pair shares a target yaw of $+60^\circ$.
Each row therefore contains two revisit comparisons:
\emph{columns 1 and 2} ($-60^\circ$), and \emph{columns 3 and 4} ($+60^\circ$).

\refstepcounter{figure}
\label{fig:app-revisit-lakeside}

\begin{appresultcase}{Case 1: Lakeside}{fig:app-revisit-lakeside-case}
    \includegraphics[page=1,trim=0 597.383 0 0.000,clip,width=\linewidth]{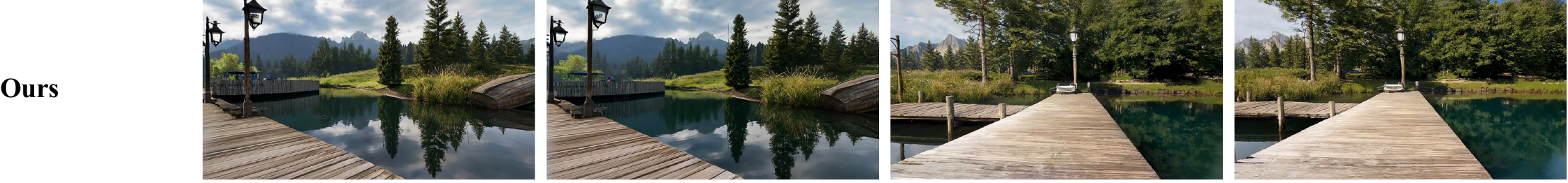}\\*
    \includegraphics[page=1,trim=0 467.765 0 125.617,clip,width=\linewidth]{assets/scene.pdf}\\
    \includegraphics[page=1,trim=0 336.742 0 255.235,clip,width=\linewidth]{assets/scene.pdf}\\
    \includegraphics[page=1,trim=0 205.718 0 386.258,clip,width=\linewidth]{assets/scene.pdf}\\*
    \includegraphics[page=1,trim=0 23.360 0 517.282,clip,width=\linewidth]{assets/scene.pdf}\\
\end{appresultcase}

\begin{appresultcase}{Case 2: Library}{fig:app-revisit-library}
    \includegraphics[page=2,trim=0 597.383 0 0.000,clip,width=\linewidth]{assets/scene.pdf}\\*
    \includegraphics[page=2,trim=0 467.765 0 125.617,clip,width=\linewidth]{assets/scene.pdf}\\
    \includegraphics[page=2,trim=0 336.742 0 255.235,clip,width=\linewidth]{assets/scene.pdf}\\
    \includegraphics[page=2,trim=0 205.718 0 386.258,clip,width=\linewidth]{assets/scene.pdf}\\*
    \includegraphics[page=2,trim=0 23.360 0 517.282,clip,width=\linewidth]{assets/scene.pdf}\\
\end{appresultcase}

\begin{appresultcase}{Case 3: Snowy ballroom}{fig:app-revisit-ballroom}
    \includegraphics[page=3,trim=0 595.976 0 0.000,clip,width=\linewidth]{assets/scene.pdf}\\*
    \includegraphics[page=3,trim=0 464.953 0 127.024,clip,width=\linewidth]{assets/scene.pdf}\\
    \includegraphics[page=3,trim=0 333.929 0 258.047,clip,width=\linewidth]{assets/scene.pdf}\\
    \includegraphics[page=3,trim=0 202.906 0 389.071,clip,width=\linewidth]{assets/scene.pdf}\\*
    \includegraphics[page=3,trim=0 20.480 0 520.094,clip,width=\linewidth]{assets/scene.pdf}\\*
    \begin{minipage}{\linewidth}
    \appfigurecaption{Static-scene revisitation across the lakeside, library,
and snowy ballroom. Columns show $5$, $10$, $20$, and $25$ seconds for every
method. Compare columns 1/2 and 3/4 for persistence of scene layout and
object appearance after returning to the same target camera pose.}
    \end{minipage}\\
\end{appresultcase}

\subsection{Long-Horizon Video Generation}

Fig.~\ref{fig:app-vbench-people} shows additional VBench
image-to-video examples from \method,
covering people, everyday activities, landscapes, and animals.

\clearpage

\refstepcounter{figure}
\label{fig:app-vbench-people}

\begin{appresultcase}{Image-to-video examples}{fig:app-vbench-cases}
    \includegraphics[page=1,trim=0 1530.748 0 20.724,clip,width=\linewidth]{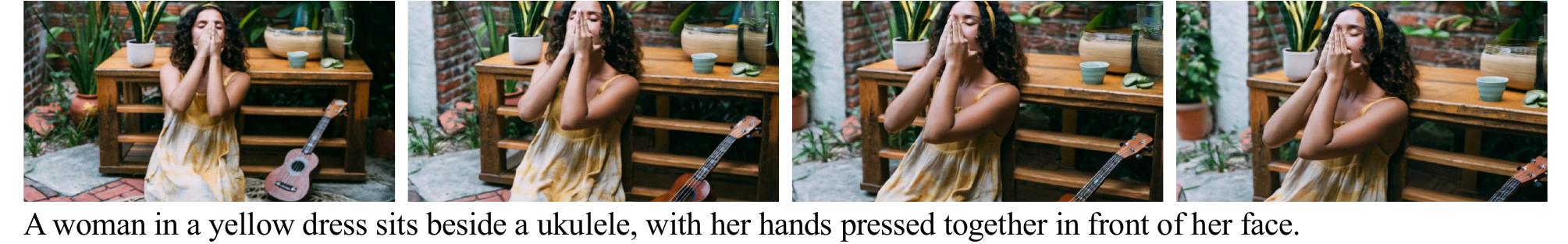}\\
    \includegraphics[page=1,trim=0 1080.666 0 470.307,clip,width=\linewidth]{assets/vbench.pdf}\\
    \includegraphics[page=1,trim=0 930.638 0 620.334,clip,width=\linewidth]{assets/vbench.pdf}\\
    \includegraphics[page=1,trim=0 771.087 0 770.362,clip,width=\linewidth]{assets/vbench.pdf}\\
    \includegraphics[page=1,trim=0 621.059 0 929.913,clip,width=\linewidth]{assets/vbench.pdf}\\
    \includegraphics[page=1,trim=0 471.031 0 1079.941,clip,width=\linewidth]{assets/vbench.pdf}\\
    \includegraphics[page=1,trim=0 321.014 0 1229.969,clip,width=\linewidth]{assets/vbench.pdf}\\
    \includegraphics[page=1,trim=0 170.998 0 1379.986,clip,width=\linewidth]{assets/vbench.pdf}\\
    \includegraphics[page=1,trim=0 21.200 0 1530.002,clip,width=\linewidth]{assets/vbench.pdf}\\*
    \begin{minipage}{\linewidth}
    \appfigurecaption{Additional results on VBench. All videos are 480p, 30s.}
    \end{minipage}\\
\end{appresultcase}
\endgroup

\end{document}